\documentclass[a4paper]{mva_style}
\usepackage{graphicx}
\usepackage{epstopdf}
\usepackage{caption}

\usepackage{enumitem} 
\usepackage{booktabs}

\usepackage{mathtools}
\DeclarePairedDelimiter\ceil{\lceil}{\rceil}
\DeclarePairedDelimiter\floor{\lfloor}{\rfloor}
\graphicspath{{Figures/}} 
\usepackage{booktabs} 
\usepackage{hyperref} 
\usepackage{tabularx}
\usepackage{color,soul} 
\soulregister\ref{7} 
\soulregister\cite{7} 
\usepackage{array} 

\usepackage{chngcntr}

\usepackage[utf8]{inputenc}
\usepackage{amsfonts}
\usepackage{url}
\usepackage{float}

\finalcopy 

\begin{document}


\title{Comparison of the Deep-Learning-Based Automated Segmentation Methods for the Head Sectioned Images of the Virtual Korean Human Project}

\author{
  Mohammad Eshghi\thanks{The first two authors contributed equally to this work.}\\
  Graduate School of Information Science\\
  Nagoya University, Nagoya, Japan\\
  {\tt eshghi@mori.m.is.nagoya-u.ac.jp}\\
  \and
  Holger R. Roth$^*$\\
  Information \& Communications\\
  Nagoya University, Nagoya, Japan\\
  {\tt rothhr@mori.m.is.nagoya-u.ac.jp}\\
  \and
  Masahiro Oda\\
  Graduate School of Information Science\\
 Nagoya University, Nagoya, Japan\\
  {\tt moda@mori.m.is.nagoya-u.ac.jp}\\
  \and
  Min Suk Chung\\
   Department of Anatomy\\ Ajou University School of Medicine\\
  Suwon, South Korea\\
  {\tt dissect@ajou.ac.kr}\\
  \and
  Kensaku Mori\\
  Information \& Communications\\Graduate School of Information Science\\
  Nagoya University, Nagoya, Japan\\
  {\tt kensaku@is.nagoya-u.ac.jp}\\
}

\maketitle


\section*{\centering Abstract}
\textit{This paper presents an end-to-end pixelwise fully automated segmentation of the head sectioned images of the Visible Korean Human (VKH) project based on Deep Convolutional Neural Networks (DCNNs). By converting classification networks into Fully Convolutional Networks (FCNs), a coarse prediction map, with smaller size than the original input image, can be created for segmentation purposes. To refine this map and to obtain a dense pixel-wise output, standard FCNs use deconvolution layers to upsample the coarse map. However, upsampling based on deconvolution increases the number of network parameters and causes loss of detail because of interpolation. On the other hand, dilated convolution is a new technique introduced recently that attempts to capture multi-scale contextual information without increasing the network parameters while keeping the resolution of the prediction maps high. We used both a standard FCN and a dilated convolution based FCN for semantic segmentation of the head sectioned images of the VKH dataset. Quantitative results showed approximately 20\% improvement in the segmentation accuracy when using FCNs with dilated convolutions.}
\section{Introduction}
\label{sec:intro}
Semantic segmentation of medical images is an  important component of many computer aided detection (CADe) and diagnosis (CADx) systems. Deep-learning-based segmentation approaches including Fully Convolutional Networks (FCN) \cite{long2015fully}, DeepLab \cite{chen2014semantic} and U-Net \cite{ronneberger2015unet}, have gained significant improvements in performance over previous methods by applying state-of-the-art CNN based image classifiers and representation to the semantic segmentation problem in both domains. Semantic segmentation involves assigning a label to each pixel in the image. Learning these dense pixel labels for each image in an end-to-end fashion is desired in many medical imaging applications. The availability of large annotated training sets and the accessibility of affordable parallel computing resources via GPUs have been paving way for segmentation based on deep learning. Systems based on deep convolutional neural networks (CNNs), like FCN, have outperformed more traditional ``shallow'' learning systems that rely on hand-crafted features. One advantage of CNNs is their build-in ability to learn features that are invariant to local image transformations. They can learn increasingly abstract layers that are useful for image classification \cite{zeiler2014visualizing,chen2016deeplab}. However, semantic segmentations tasks might suffer from this increased invariance to local transformations where dense prediction results are required. Furthermore, the combination of max-pooling and downsampling layers in CNNs decrease the spatial resolution of the feature space which make dense prediction at the full image resolution difficult \cite{long2015fully}. Recently, Wang et al. \cite{chen2016deeplab} addressed these issues when applying CNNs for semantic image segmentation. In order to produce denser feature maps, downsampling layers are removed from the last few max pooling layers and instead introduce multi-scale filters in the subsequent convolutional layers \cite{chen2016deeplab}. The multi-scale filters are realized as `dilated convolution' layers that allow the feature maps to be computed at a higher sampling rate. Dilated convolutions effectively enlarge the field of view without increasing the number of parameters or the amount of computation \cite{chen2016deeplab}. Dilated convolutions can be used to resample a given feature layer at multiple rates during convolution. This effectively allows the CNN to compute features at different scales of the input image, similar in spirit to spatial pyramid pooling \cite{he2014spatial}. \\
\indent While standard FCNs have been widely applied to the biomedical imaging field \cite{wang2016deep,roth2016spatial,bentaieb2016topology,xu2016gland,ben2016fully}, CNNs employing dilated convolutions have not yet been well studied. In this study we compare an off-the-shelf CNN with dilated convolutions (DeepLabv2 \cite{chen2016deeplab}) with the standard FCN \cite{long2015fully} and show its advantage to the task of semantic segmentation in biomedical imaging.\\
\indent The rest of this work is structured as follows. In section \ref{sec:method}, we briefly present standard FCNs \cite{long2015fully} and dilated-convolution-based FCNs for semantic segmentation. Experiments will be addressed in section  \ref{sec:results}. Section \ref{sec:discussion} includes discussion. Summary and conclusion can be found in section \ref{sec:conclusion}.
\section{Method}
\label{sec:method}
\subsection{Standard fully convolutional networks for semantic segmentation}
In end-to-end semantic segmentation, the idea is to directly predict a label for each pixel in the input image. To achieve a dense and pixel-to-pixel label prediction, one must integrate the local pixel-level information with the wider global context information.

Existing state-of-the-art networks for semantic segmentation based on fully convolutional networks \cite{long2015fully} are typically designed based on integration of multi-scale contextual information, relying on successive spatial pooling and subsampling \cite{Yu2016multi}, to obtain a prediction. Due to the fact that both pooling and convolution reduce the spatial extent of the feature maps, additional unpooling and deconvolution (including bilinear upsampling) layers are required to make a final end-to-end pixelwise prediction. 
\subsection{Dilated convolution and semantic segmentation}
The drawback of using deconvolution layers is that they increase the number of parameters (weights) in the network. To resolve this issue, \cite{Yu2016multi} and \cite{chen2016deeplab} have recently developed a new convolutional network module based on dilated convolution (also known as `\textit{atrous}' convolution), which can compute the responses of various layers without any loss in spatial resolution.\\
\indent Let $I_{in} \in \mathbb{R}^{P\times Q}$, $k \in \mathbb{R}^{M\times N}$ and $I_{out} \in \mathbb{R}^{P\times Q}$ be input image, arbitrary discrete filter kernel and output image, respectively. Further let $r \in \mathbb{N}$ be convolution rate or dilation factor, with $\mathbb{N}$ being the set of natural numbers. The discrete  $r$-dilated convolution in 2D is then defined as \cite{chen2016deeplab}
\begin{equation} \label{dilConv}
\begin{split}
I_{out}(i, j) & = [I_{in}\ast_r k] (i,j)\\
& =\hspace{-0.3cm} \sum_{m =  \ceil*{\frac{-M}{2}}}^{\floor*{\frac{M}{2}}}\sum_{n =  \ceil*{\frac{-N}{2}}}^{\floor*{\frac{N}{2}}}\hspace{-0.3cm}I_{in}(i + r\cdot m,\ j+r\cdot n)k(m,n)\\
& =\hspace{-0.3cm} \sum_{m =  \ceil*{\frac{-M}{2}}}^{\floor*{\frac{M}{2}}}\sum_{n =  \ceil*{\frac{-N}{2}}}^{\floor*{\frac{N}{2}}}\hspace{-0.3cm}k(i + r\cdot m,\ j+r\cdot n)I_{in}(m,n),\\
\end{split}
\end{equation}
where $[\cdot\ast \cdot]$, $\ceil{\cdot}$ and $\floor{\cdot}$ denote discrete convolution, ceil and floor operators respectively. Here we set $P = Q$ and $M=N$, to achieve both square input images and square filter kernels. Note that Eq. (\ref{dilConv}) is a generalized definition of the 2D discrete convolution (this can be verified easily by setting the dilation factor $r$ to 1). 

\begin{figure}[H]
  \begin{minipage}[b]{1\linewidth}
    \centering
     \centerline{\includegraphics[trim = 0mm 160mm 30mm 22mm, clip,width=0.75\textwidth]{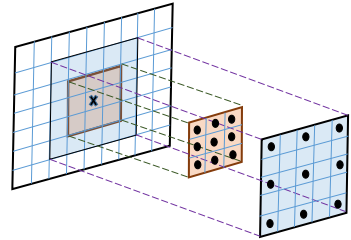}}
\end{minipage}
    \caption{Dilated Convolution.}
  \label{fig:DilatedFilter}
\end{figure}

The advantage of using dilated convolutions is that they can be considered as convolution of the original image with the filter kernel, upsampled by a factor $r$, hence they increase the receptive fields of the neurons without losing spatial resolution. More precisely, during the upsampling of the kernel, we are effectively appending some zeros in between filter values (see Fig. \ref{fig:DilatedFilter}). 

\section{Experiments}
\label{sec:results}
\begin{figure}[H]
\begin{minipage}[b]{1\linewidth}
  \centering
    \centerline{\includegraphics[trim = 0mm 0mm 80mm 0mm, clip,width=1.1\textwidth]{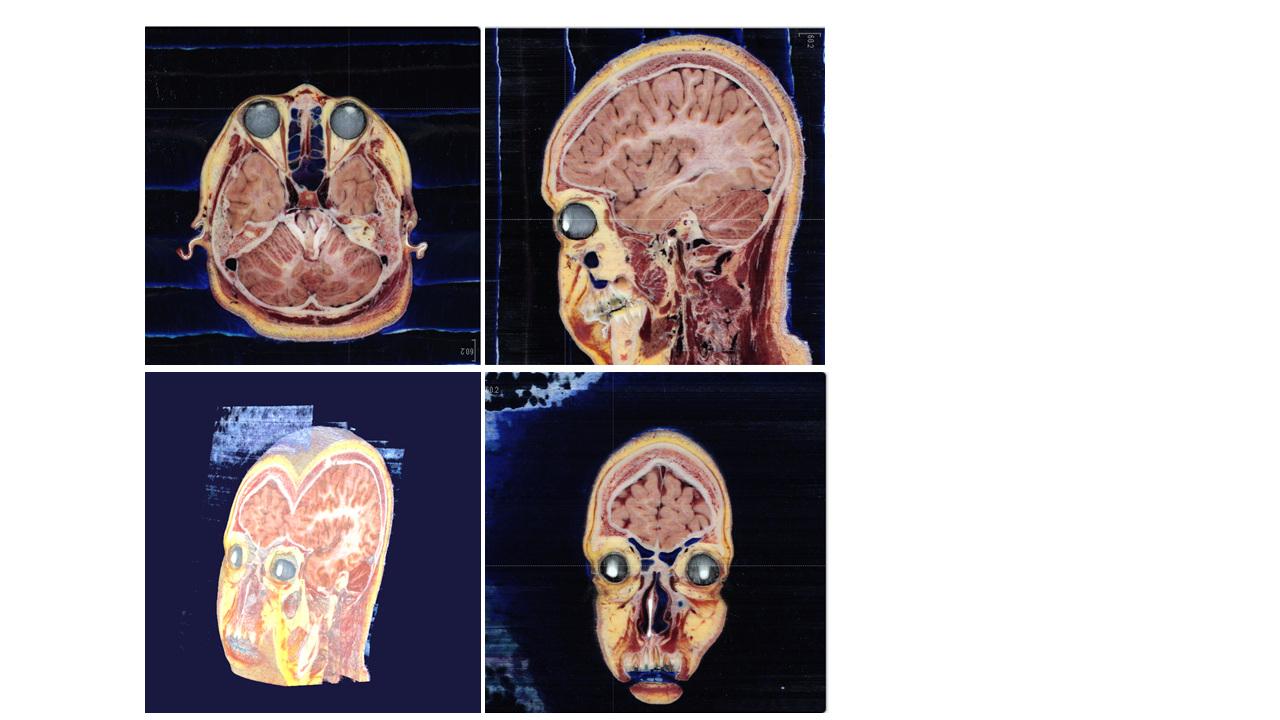}}
  \end{minipage}
\caption{A typical cross-section of the (VKH) dataset. The 3D volume in the bottom left corner has been rendered by VAA3D \cite{vaa3d}.}
\label{fig:originalImg}
\end{figure}

\textbf{Data:} For our experiments, we selected sectioned images of the head from of the \textbf{V}isible \textbf{K}orean \textbf{H}uman (VKH) dataset of the male cadaver. This dataset has been created by Prof. Min Suk Chungin, Department of Anatomy, Ajou University School of Medicine, Suwon, South Korea. In this dataset, the sectioned anatomical images have been photographed using a digital camera, Canon EOS 5D, with 12 mega pixels resolution and 0.1 mm pixel size, and they have been stored as 5616$\times$2300 color images (see \cite{VKPhome} for more information). We cropped all images to a size of 1024$\times$1024 that covers the entire head region. A typical cross-section of the VKH dataset is shown in Fig. \ref{fig:originalImg}. Manual segmentation of each cross-sectional slice was performed in PLUTO\footnote{\url{http://pluto.newves.org/trac}} in order to label 8 regions, including background, skull, teeth, cerebrum, cerebellum, nasal cavities, eyeballs, and lenses. 

\textbf{Experiments:} We investigated the following three use cases of FCNs and dilated convolution based FCNs:

\textit{1) Performance comparison of standard FCN vs. dilated convolution based FCN:} to compare the resulting segmentation accuracy and to show the advantage of utilizing dilated convolution in FCNs, we conducted an experiment in which a random subset of 80\% of the images was used for training, while 20\% of the images were reserved for testing the networks' performance. 
\\
\indent\textit{2) Label propagation based on sparse annotation:} the basic idea here is that we are interested in labeling a random subset of the slices to be considered as ground truth (sparse annotation), and let the labels propagate through the whole remaining slices in the dataset by the trained network (label propagation).  To this end, in the second experiment we swapped the related percentages of the slices for training and testing (20$\%$ for training, 80$\%$ for testing).
\\
\indent\textit{3) Generalizability capability:} to show the generalizability of the trained network, in the third experiment we applied the trained DeepLabv2 model (trained on 80$\%$ of the slices from the dataset introduced in section \ref{sec:results}) to another unseen VKH dataset, for which no ground truth was available, and we aimed to qualitatively evaluate the performance of the network.

\textbf{Implementation:} All experiments were conducted on a workstation equipped with one NVIDIA GeFORCE graphic cards, NVIDIA GeForce GTX 1080,  and two 3.20 GHz Intel Xeon X5482  processors with a 64-bit Ubuntu 14.04 and 32 GB RAM memory. We used Caffe implementations \cite{jia2014caffe} of FCN\footnote{\url{https://github.com/shelhamer/fcn.berkeleyvision.org}} and DeepLabv2\footnote{\url{https://bitbucket.org/aquariusjay/deeplab-public-ver2}}.

\textbf{Evaluation:} 
We evaluated our results for the first two experiments both qualitatively and quantitatively. For the third experiment, lack of ground truth, only qualitative evaluation was performed. Networks' performance for quantitative evaluation was measured based on Dice Similarity Coefficient (DSC).
\section{Discussion}
\label{sec:discussion}
All experiments were conducted on 2D RGB images. Figure \ref{fig:resultComp} illustrates the achieved fully automated segmentation results for the given cross-sectional images shown in Fig. \ref{fig:originalImg}. This figure shows that FCNs based on dilated convolution could obtain smoother segmentation results with lower false-positive rate (higher accuracy) than the standard FCNs. \\ 
\indent In terms of numbers, the quantitative evaluation results have been summarized in Table \ref{tab:dice}. To show the advantage of utilizing dilated convolution in FCNs, for every individual label in the ground truth, the corresponding DSC values both for training and testing phases have been calculated. Considering the mean and standard deviation values over all labels especially in the testing phase and with p-value or significance level less than 0.01 for Wilcoxon signed rank test, it is evident that by using dilated convolution the increase in testing DSC performance (\textbf{\text{\boldmath$\Delta$}test}) is significant (here 19.6\% on average, as in Table \ref{tab:dice}), whereas at the same time the standard deviation has been decreased by 11.2\%. This indicates that the overall segmentation accuracy of the network has been improved. The increased contextual information used by DeepLabv2 is clearly helping the network to achieve more coherent and less noisy results.\\
%
%
%
\indent In the second experiment we swapped the related percentages of the slices for training and testing (20$\%$ for training, 80$\%$ for testing). Interestingly, the network could achieve quite the same DSC values, as in the case with 80$\%$ of the slices for training. Quantitative and qualitative results for label propagation can be found in the last column of Table \ref{tab:dice} and in Fig \ref{fig:applications}-(a), respectively.\\
\indent Another important issue to mention here is that the labeling process of anatomical dataset is in general a tedious and time-consuming task. In terms of practical applications, it would be of particular interest if the labeling process, which has been done for one dataset, could be generalized to other similar dataset. Our results from the third experiment showed that the network was able to achieve comparable segmentation results as shown in Fig \ref{fig:applications}-(b).

\begin{table}[t]
\vspace{0.5cm}
  \caption{Dice Similarity Coefficient (DSC) in testing in comparison between FCN and DeepLabv2. The advantage of using dilated convolutions in DeepLabv2 is clearly visible in the $ \textbf{\text{\boldmath$\Delta$}test}$ values ($\Delta$ denotes the difference between DeepLab and FCN results).}
\begin{minipage}[b]{0.1\linewidth}
  \centering
\resizebox{10\columnwidth}{!}{
\setlength{\extrarowheight}{20pt}
  \begin{tabular}{*{7}{|c|c|c|c|c|c|c}}
\specialrule{1.5pt}{1pt}{1pt}
   \Huge \textbf{Class} &\Huge \textbf{Train} &\Huge \textbf{Test} &\Huge \textbf{Train-Deep} &\Huge \textbf{Test-Deep} &\Huge \textbf{\text{\boldmath$\Delta$}test} & \Huge \textbf{Test-Deep}\\[4pt] 
   \Huge \textbf{} &\Huge \textbf{FCN-80$\%$} &\Huge \textbf{FCN-20$\%$} &\Huge \textbf{Labv2-80$\%$} &\Huge \textbf{Labv2-20$\%$} &\Huge \textbf{} & \Huge \textbf{Labv2-80$\%$}\\ [4pt]
\specialrule{2pt}{1pt}{1pt}
\Huge Background &\Huge 98.6\% &\Huge 98.1\% &\Huge 99.6\% &\Huge 99.6\%  &\Huge 1.5\%  &\Huge 99.6\%\\ [4pt]
    \hline 
\Huge Skull &\Huge 80.7\% &\Huge 71.6\% & \Huge 93.7\% &\Huge 93.0\% &\Huge 21.4\% &\Huge 99.3\% \\[4pt]
    \hline
\Huge Teeth &\Huge 75.1\% &\Huge 52.6\% &\Huge 75.4\% &\Huge 74.3\%  &\huge 21.7\% &\Huge 74.7\% \\[4pt]
    \hline
\Huge Cerebrum &\Huge 95.3\% &\Huge 92.2\% &\Huge 98.9\% &\Huge 98.8\%  &\Huge 6.6\% &\Huge 98.8\%\\[4pt]
    \hline
\Huge Cerebellum &\Huge 78.7\% &\Huge 73.6\%&\Huge 97.6\% &\Huge 97.4\%  &\Huge 23.8\% &\Huge 96.6\%\\[4pt]
    \hline
\Huge Nasal Cavities &\Huge 60.3\% &\Huge 55.4\% &\Huge 88.2\% &\Huge 88.7\%  &\Huge 33.3\% &\Huge 88.7\%\\[4pt]
    \hline
\Huge Eyeballs &\Huge 91.7\% &\Huge 77.9\% &\Huge 94.1\% &\Huge 93.9\% &\Huge 16.0\% &\Huge 93.5\% \\[4pt]
    \hline
\Huge Lenses &\Huge 76.4\% &\Huge 46.6\% &\Huge 79.9\% &\Huge 78.9\% &\Huge 32.3\% &\Huge 77.2\%\\
\specialrule{2pt}{1pt}{1pt}
\Huge\textbf{Mean} &\Huge $\mathbf{82.1}\textbf{\%}$ &\Huge $\mathbf{71.0}\textbf{\%}$ &\Huge $\mathbf{90.9}\textbf{\%}$ &\Huge $\mathbf{90.6}\textbf{\%}$ &\Huge $\mathbf{19.6}\textbf{\%}$ &\Huge $\mathbf{90.2}\textbf{\%}$\\[4pt]
    \hline
\Huge \textbf{Std. dev.} &\Huge $\mathbf{12.6}\textbf{\%}$ &\Huge $\mathbf{18.6}\textbf{\%}$ &\Huge $\mathbf{9.0}\textbf{\%}$ &\Huge $\mathbf{9.4}\textbf{\%}$  &\Huge $\mathbf{11.2}\textbf{\%}$ &\Huge $\mathbf{9.5}\textbf{\%}$\\[4pt]
    \hline
\Huge \textbf{Min} &\Huge $\mathbf{60.3}\textbf{\%}$ &\Huge $\mathbf{46.6}\textbf{\%}$ &\Huge $\mathbf{75.4}\textbf{\%}$ &\Huge $\mathbf{74.3}\textbf{\%}$  &\Huge $\mathbf{1.5}\textbf{\%}$&\Huge $\mathbf{74.7}\textbf{\%}$ \\[4pt]
 \hline
\Huge \textbf{Max} &\Huge $\mathbf{98.6}\textbf{\%}$ &\Huge $\mathbf{98.1}\textbf{\%}$ &\Huge $\mathbf{99.6}\textbf{\%}$ &\Huge $\mathbf{99.6}\textbf{\%}$ &\Huge $\mathbf{33.3}\textbf{\%}$ &\Huge $\mathbf{99.6}\textbf{\%}$\\[4pt]
\specialrule{2pt}{1pt}{1pt}
  \end{tabular}}
\end{minipage}
  \label{tab:dice}
\end{table}
\begin{figure}[t]
\begin{minipage}[b]{.495\linewidth}
\centering
\centerline{\includegraphics[trim = 60mm 0mm 300mm 5mm, clip,width=1\textwidth]{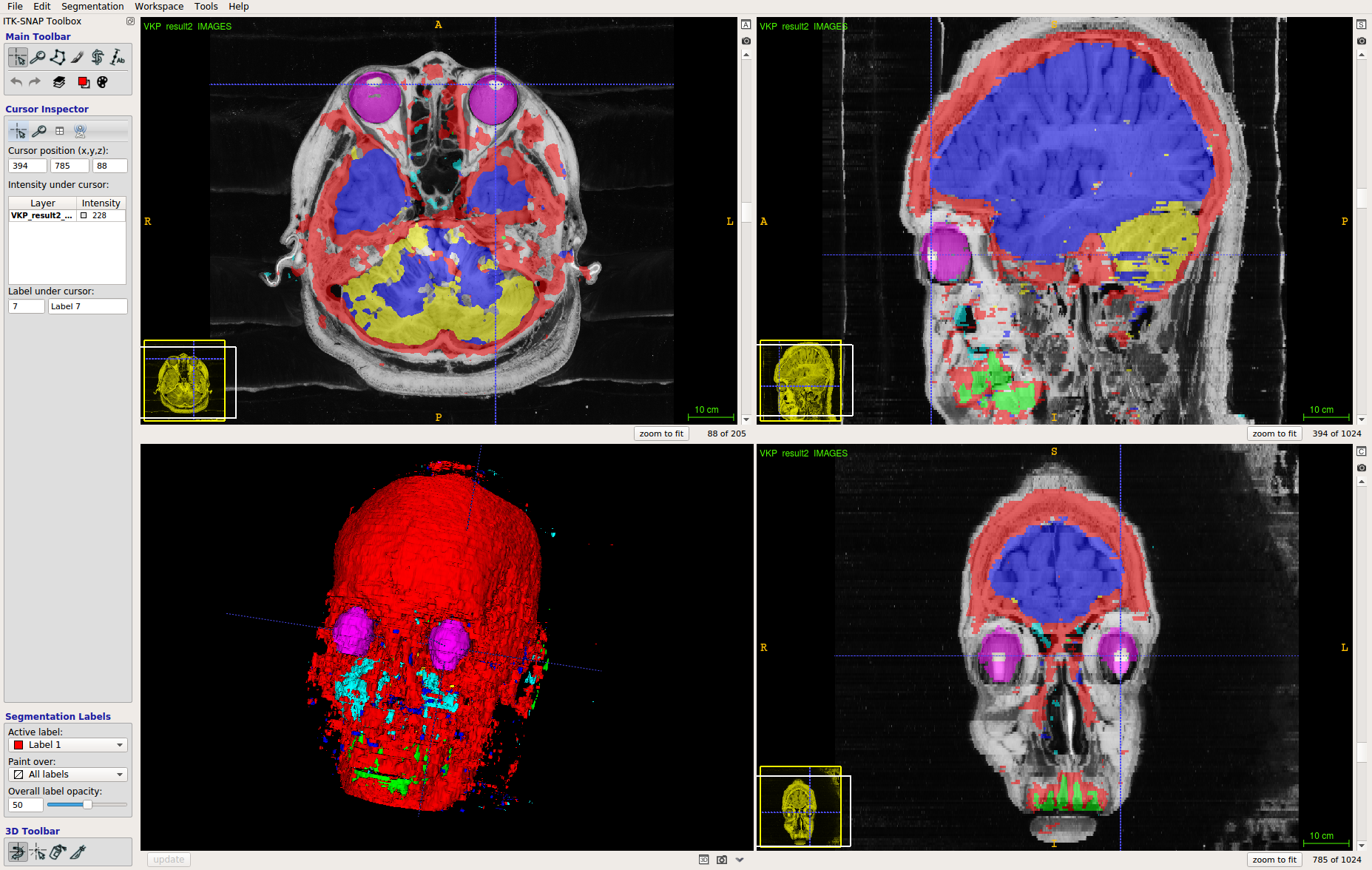}}
\vspace{.1cm}
\centerline{(a) Standard FCN}\medskip
\end{minipage}
\hfill
\begin{minipage}[b]{.495\linewidth}
\centering
\centerline{\includegraphics[trim = 60mm 0mm 300mm 5mm, clip,width=1\textwidth]{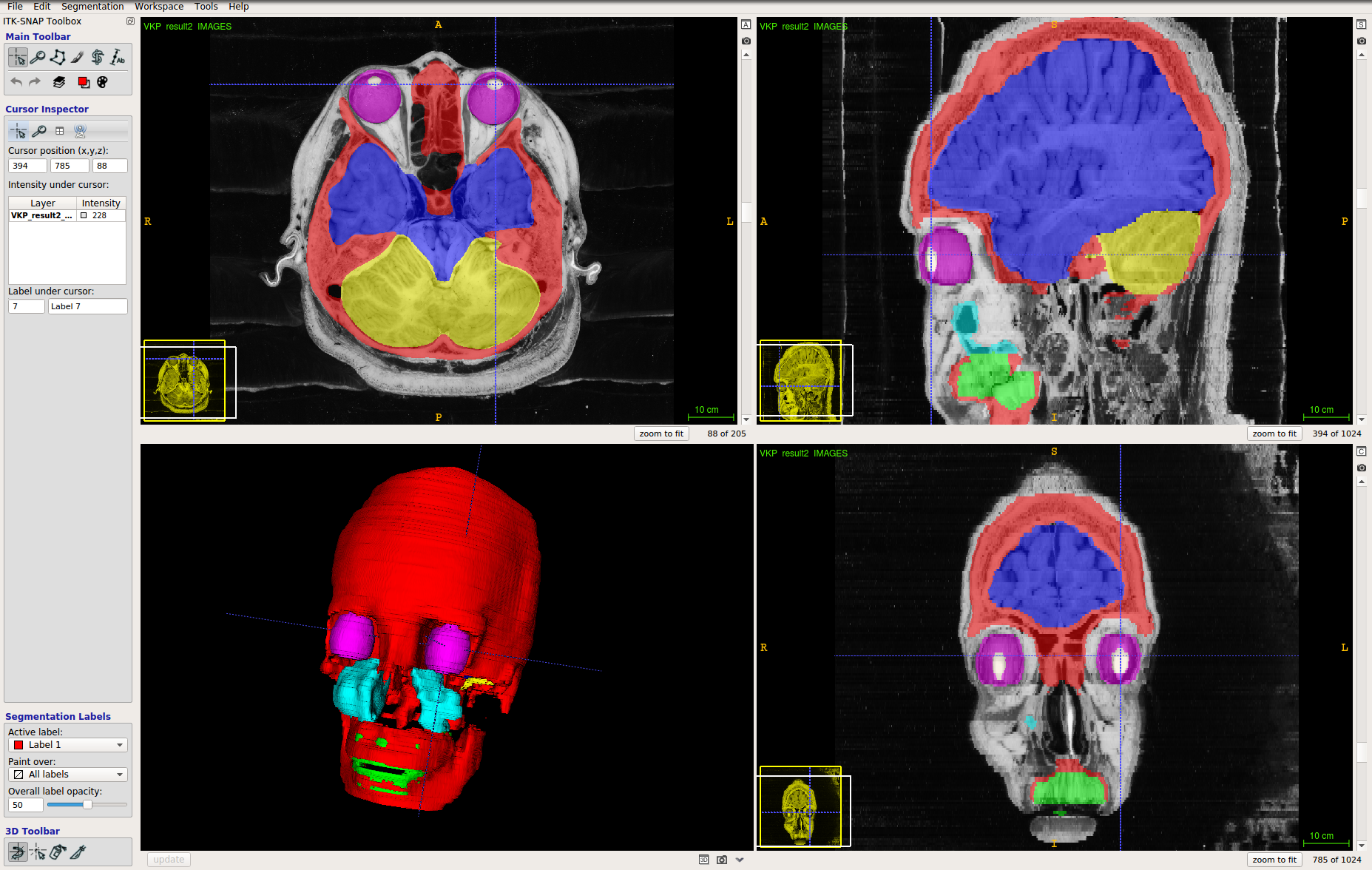}}
\vspace{.1cm}
\centerline{(b) DeepLabv2}\medskip
\end{minipage}
\vspace{-0.4cm}
\caption{Comparison of the segmentation results.}
\label{fig:resultComp}
\end{figure}

%

\begin{figure*}[t]
\vspace{3mm}
\begin{minipage}[b]{.5\linewidth}
\centering
\centerline{\hspace{-0.2cm}\includegraphics [trim = 100mm 20mm 84mm 55mm, clip, scale = 0.5, width = 0.91\linewidth ]{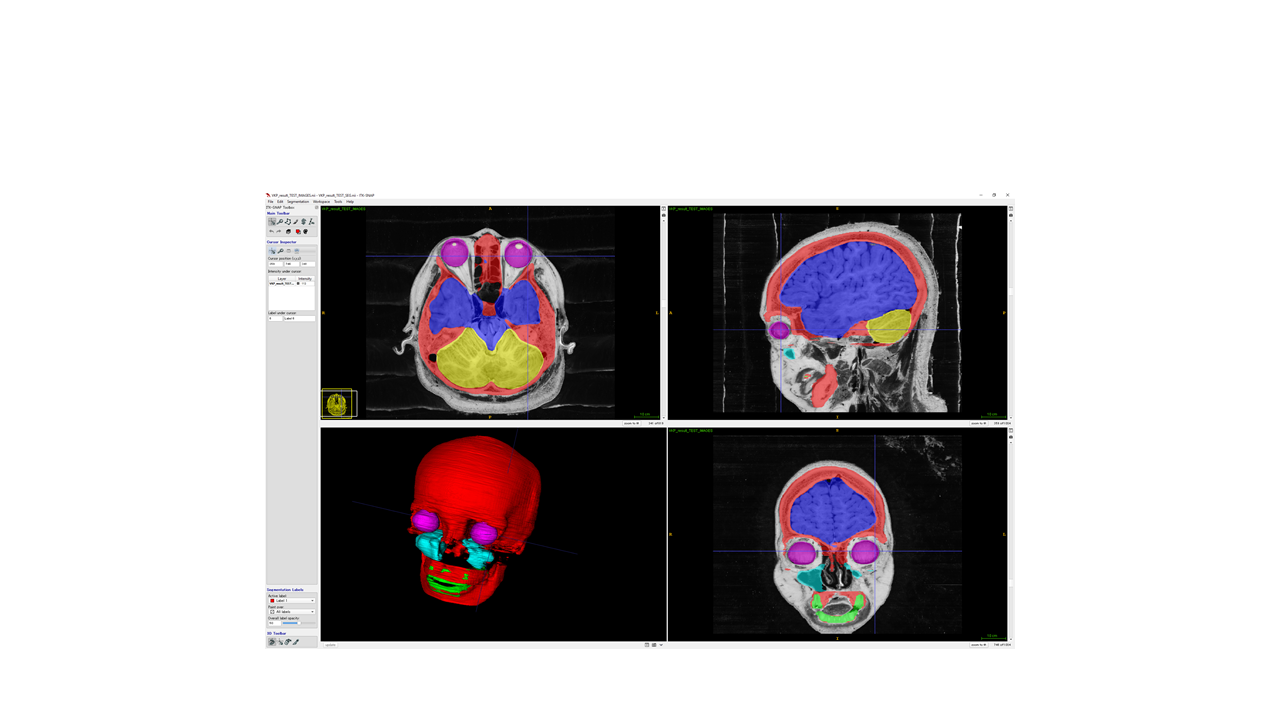}}
\captionsetup{labelformat=empty, width=0.9\textwidth}
\caption{(a) Sparse annotation (training based on 20\% of the slices), and the resulted labels propagation (testing for the remaining 80\% of the slices).}\medskip
\end{minipage}
\hfill
\begin{minipage}[b]{.5\linewidth}
\centering

\centerline{\hspace*{0.1cm} \raisebox{-2mm}[0pt][0pt]{\includegraphics[trim = 20mm 5mm 6mm 0mm, clip, width = 0.92\linewidth]{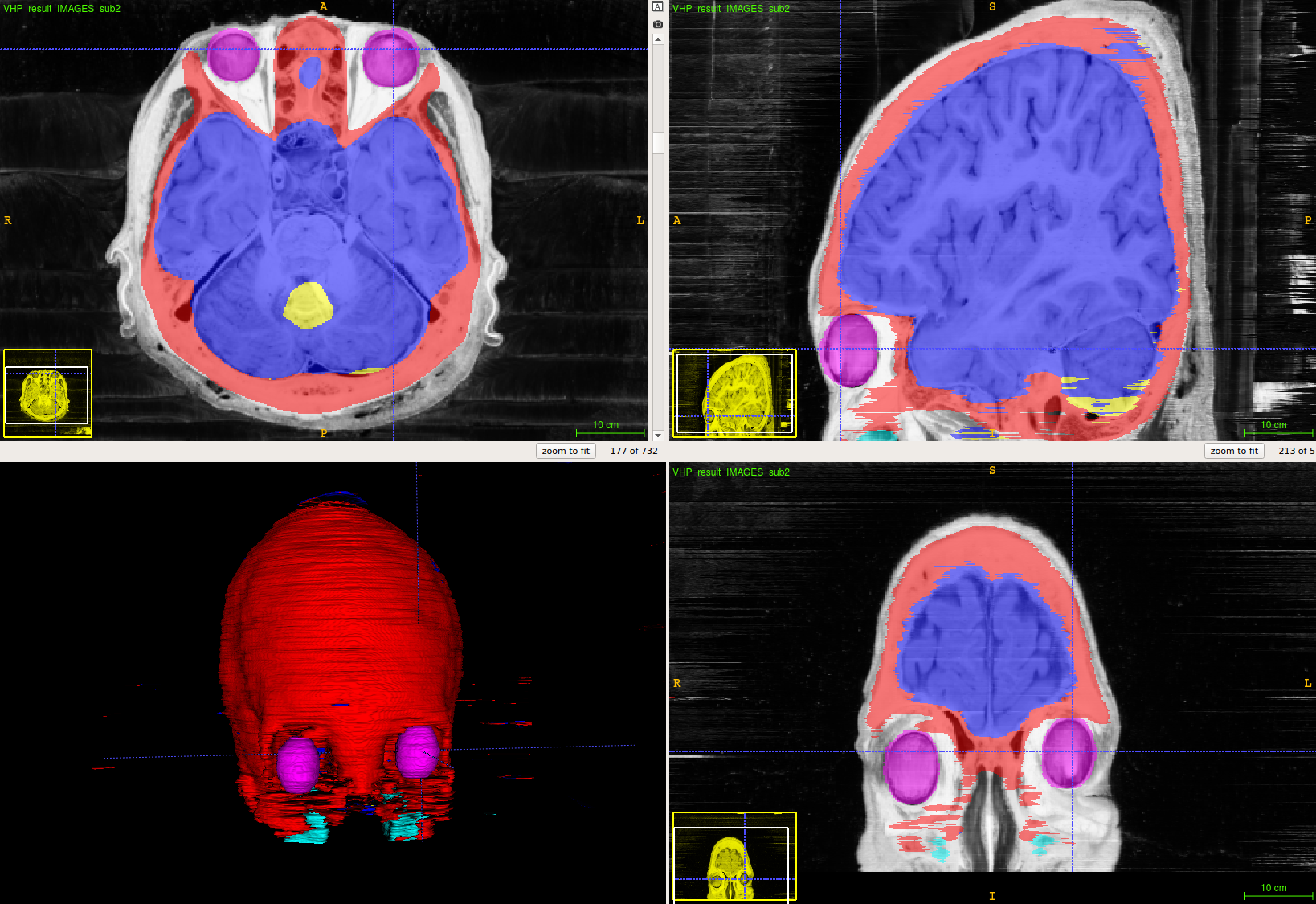}}}
\captionsetup{labelformat=empty, width=0.9\textwidth}
\caption{(b) Generalizability of the trained network: the DeepLabv2 network was trained on the dataset explained in section \ref{sec:results}, and it was used for segmenting the same labels in this unseen dataset.}\medskip
\end{minipage}
\setcounter{figure}{3}
\caption{Practical applications of the dilated-convolution-based trained network.}
\label{fig:applications}
\end{figure*}

\section{Summary and Conclusion}
\label{sec:conclusion}
We provided experimental results that show the advantage of using dilated convolution in deep fully convolutional architectures. Utilizing dilated convolutions allows the increase of the DCNN's receptive fields while keeping the resolution of feature maps high, allowing for denser semantic segmentation results at the final layers. We investigated the feasibility of the label propagation based on sparsely-trained model, and the generalizability of the network for segmenting an unseen dataset. Training and quantitative testing on the VKH dataset shows the applicability of these methods for biomedical imaging.

\section*{Acknowledgment}
\noindent Part of this work is supported by the ImPACT and the JSPS KAKENHI (Grant Numbers 26108006, 26560255, and 25242047).

%
%
%

%
%
%
%

\bibliographystyle{ieeetr} 
\bibliography{references_MVA2017}

\end{document}